\documentclass{article} 
\pdfoutput=1
\usepackage{times}
\usepackage{url}
\usepackage{amsmath, amsthm, amssymb}
\theoremstyle{plain}
\theoremstyle{definition}

\usepackage[T1]{fontenc}
\usepackage{subfigure}
\usepackage{graphicx}
\usepackage{caption}
\usepackage{amsfonts}
\usepackage{algorithm}
\usepackage{algcompatible}
\usepackage[font=footnotesize]{caption}
\usepackage{subfigure}
\usepackage{multicol}
\usepackage{ifthen}
\usepackage[numbers,sort&compress]{natbib}
\usepackage[bookmarks=true]{hyperref}

\newcommand{\nt}{\\ \notag}

\newcommand*{\meq}[2]{\begin{align}#2\begin{matrix}#1\end{matrix}\end{align}}

\newcommand{\eref}[1]{(\ref{#1})}
\newcommand{\aref}[1]{Algorithm~\ref{#1}}
\newcommand{\sref}[1]{Section~\ref{#1}}
\newcommand{\figref}[1]{Figure~\ref{#1}}

\newcommand{\bs}[1]{\boldsymbol{#1}}

\newboolean{include-notes}
\setboolean{include-notes}{true}
\newcommand{\ssnote}[1]{\ifthenelse{\boolean{include-notes}}%
 {\textcolor{orange}{\textbf{[ #1  --Sidd]}}}{}}
\newcommand{\adnote}[1]{\ifthenelse{\boolean{include-notes}}%
 {\textcolor{LimeGreen}{\textbf{[ #1  --Anca]}}}{}}
 \newcommand{\gnote}[1]{\ifthenelse{\boolean{include-notes}}%
 {\textcolor{purple}{\textbf{[ #1 --Geoff]}}}{}}
  \newcommand{\bnote}[1]{\ifthenelse{\boolean{include-notes}}%
 {\textcolor{green}{\textbf{[ #1  --Byron]}}}{}}
 \newcommand{\abnote}[1]{\ifthenelse{\boolean{include-notes}}%
 {\textcolor{blue}{\textbf{[#1  --Arun]}}}{}}
  \newcommand{\znote}[1]{\ifthenelse{\boolean{include-notes}}%
 {\textcolor{cyan}{\textbf{[ #1  --Zita]}}}{}}
\hypersetup{bookmarksopen,bookmarksnumbered,
pdfpagemode=UseOutlines,
colorlinks=true,
linkcolor=blue,
anchorcolor=blue,
citecolor=blue,
filecolor=blue,
menucolor=blue,
urlcolor=blue,
}

\begin{document}

\title{Functional Gradient Motion Planning\\ in Reproducing Kernel Hilbert Spaces}

\author{}


\author{
Z. Marinho\\
Robotics Institute, CMU\\
\texttt{zmarinho@cmu.edu} \\
\and
A. Dragan \\
Robotics Institute, CMU \\
\texttt{adragan@cs.cmu.edu} \\
\and
A. Byravan \\
University of Washington \\
\texttt{barun@uw.edu} \\
\and
B. Boots \\
Georgia Institute of Technology \\
\texttt{bboots@cc.gatech.edu} \\
\and
G. Gordon \\
 Machine Learning Dep., CMU \\
\texttt{ggordon@cs.cmu.edu} \\
\and
S. Srinivasa\\
Robotics Institute, CMU \\
\texttt{siddh@cs.cmu.edu} \\
}

\maketitle

\begin{abstract}
We introduce a functional gradient descent trajectory optimization algorithm
for robot motion planning in Reproducing Kernel Hilbert Spaces
(RKHSs). Functional gradient algorithms are a popular choice for motion planning
in complex many-degree-of-freedom robots, since they (in theory) work
by directly optimizing within a space of continuous trajectories to avoid obstacles
while maintaining geometric properties such as 
smoothness. However, in practice, functional gradient algorithms typically
\emph{commit to a fixed, finite parameterization}
 of trajectories, often as a list of waypoints. Such a
parameterization can lose much of the benefit of reasoning in a
continuous trajectory space: e.g., it can
require taking an inconveniently small step size and large number of
iterations to maintain smoothness. 
Our work generalizes functional gradient trajectory optimization by formulating it as minimization
of a cost functional in an RKHS\@.  
This generalization lets us represent trajectories as linear combinations of
kernel functions, without any need for waypoints.  As a result, we are able to take larger
steps and achieve a locally optimal trajectory in just a few iterations. Depending
on the selection of kernel, we can directly optimize in spaces of trajectories
that are \emph{inherently smooth} in velocity, jerk, curvature, etc., and that have a \emph{low-dimensional}, adaptively
chosen parameterization. Our experiments illustrate the effectiveness of the planner
for different kernels, including Gaussian RBFs, Laplacian RBFs, and B-splines, as compared to the standard
discretized waypoint representation.
\end{abstract}

%
%
\section{Introduction \& Related Work}
\label{sec:intro}
Motion planning is an important component of robotics: it ensures that robots are able to safely move from a start to a goal configuration without colliding with obstacles. \emph{Trajectory optimizers} for motion planning focus on finding feasible configuration-space trajectories that are also {efficient}---e.g., approximately locally optimal for some cost function.
Recently, trajectory optimizers have demonstrated great success in a number of high-dimensional real-world problems.~\cite{quinlan1993elastic, schulman2013finding, todorov2005generalized, van2011lqg}  Often, they work by defining a cost functional over an infinite-dimensional Hilbert space of trajectories, then taking steps down the functional gradient of cost to search for smooth, collision-free trajectories.~\cite{Zucker13, Ratliff09}
In this work we exploit the same functional gradient approach, but with a novel approach to trajectory representation. While previous algorithms are derived for trajectories in Hilbert spaces in theory, \emph{in practice} they commit to a finite parameterization of trajectories in order to instantiate a gradient update~\cite{Zucker13,park2012itomp,stomp}---typically a large but finite list of discretized waypoints. The number of waypoints is a parameter that trades off between computational complexity and trajectory expressiveness. 
Our work frees the optimizer from a discrete parameterization, enabling it to perform gradient descent on a much more general trajectory parameterization: reproducing-kernel Hilbert spaces (RKHSs),~\cite{smolabook,KW71,aronszajn50} of which waypoint parameterizations are merely one instance. RKHSs impose just enough structure on generic Hilbert spaces to enable a concrete and implementable gradient update rule, while leaving the choice of parameterization flexible: different kernels lead to different geometries.

Our contribution is two-fold. Our \emph{theoretical} contribution is the formulation of functional gradient descent motion planning in RKHSs, as the minimization of a cost functional regularized by the RKHS norm.  
Regularizing by the RKHS norm is a common way to ensure smoothness in function approximation,~\cite{hofman08} and we apply the same idea to trajectory parametrization.  By choosing the RKHS appropriately, the trajectory norm can quantify different forms of smoothness or efficiency, such as any low $n$-th order derivative.~\cite{yuan10} So, RKHS norm regularization can be tuned to prefer trajectories that are smooth with, for example, low velocity, acceleration, or jerk. 

Our \emph{practical} contribution is an algorithm for very efficient motion planning in inherently smooth trajectory space with low-dimensional parameterizations. Unlike discretized parameterizations, which require many waypoints to produce smooth trajectories, our algorithm can represent and search for smooth trajectories with only a few point evaluations. 
The inherent smoothness of our trajectory space also increases efficiency; our parametrization allows the optimizer to take large steps at every iteration without violating trajectory smoothness, therefore converging to a collision-free trajectory faster than competing approaches.

Our experiments demonstrate the effectiveness of planning under RKHS, and show how different choices of kernels yield different forms of trajectory efficiency.
\sref{sec:results} illustrates these advantages of RKHSs, and compares different choices of kernels. 


\section{Trajectories in an RKHS}
\label{sec:planning_rkhs}
In this paper we perform trajectory optimization in a more restricted space of trajectories, we constrain the domain where trajectories are defined to \emph{Reproducing Kernel Hilbert Spaces}. We trade off representational power for an inherent smooth representation of trajectories, given by a kernel metric. 

A trajectory is a function $\boldsymbol\xi:[0,1]\to\mathcal C$ mapping time $t\in[0,1]$ to robot configurations $\boldsymbol{\xi}(t)\in\mathcal{C}\equiv\mathbb R^D$. We can treat a set of trajectories as a Hilbert space by defining vector-space operations such as addition and scalar multiplication of trajectories.~\cite{kreyszig92}  And, we can upgrade our Hilbert space to an RKHS $\mathcal H$ by assuming additional structure: for any $y\in\mathcal C$ and $t\in[0,1]$, the functional $\xi\mapsto\langle y, \xi(t)\rangle$ must be continuous.~\cite{smolabook,W98,Ratliff07}
Note that, since configuration space is typically multidimensional ($D>1$), our trajectories form an RKHS of \emph{vector-valued} functions,~\cite{pontil05} defined by the above property.
The reproducing Kernel associated with a vector valued RKHS, becomes a matrix valued kernel ${K}: [0,1]\times[0,1]\rightarrow \mathcal{C}\times\mathcal{C}$. Eq.~\ref{eq:vecrkhs} represents the kernel matrix of joint interactions for two different time instances:
\begin{align}\label{eq:vecrkhs}
{K}(t,t')=\begin{bmatrix}
k_{1,1}(t,t')& k_{1,2}(t,t')\ \  \dots\  k_{1,D}(t,t')\\
k_{2,1}(t,t')& k_{2,2}(t,t')\ \  \dots\  k_{2,D}(t,t')\\
\vdots &\ddots\ \ \ \ \ \ \ \ \ \ \ \vdots\\
k_{D,1}(t,t')& k_{D,2}(t,t')\ \  \dots\ k_{D,D}(t,t')
\end{bmatrix}
\end{align}
This matrix has a very intuitive physical interpretation. It can be regarded as an inertia tensor of a rigid body changing over time.
Each element $k_{d,d'}(t,t')$ tells us how joint $\xi_d(t)$ affects the motion of joint $\xi_{d'}(t')$, \emph{i.e.} its degree of correlation or similarity between the two configurations.
In practice, off-diagonal terms of \eref{eq:vecrkhs} will not be zero, hence perturbations of a given joint $d$ propagate through time, as well as, through the rest of the joints. 
The norm and inner product defined in a coupled RKHS can be written in terms of the kernel matrix, via the reproducing property (trajectory evaluation can be represented as an inner product of the vector valued functions in the RKHS):
\meq{}{
\bs{y}^{\top}\bs{\xi}(\cdot)&= \langle \xi, {K}(t,\cdot) \bs{y}\rangle_{\mathcal{H}},\ \ \forall{\bs{y}\in \mathcal{C}}
}

A trajectory in the RKHS admits a representation in terms of the finite support \{$t_i\}_{i=1}^{N}\in \mathcal{T}$.
\meq{}{
\bs{y}^{\top}\bs{\xi}^{\ast} (\cdot)&= \sum_{t_i\in \mathcal{T}} a_{i} {K}(t,t_i)\bs{y}}
If we consider the configuration vector $\bs{y}\equiv \bs{e}_d$ to be the indicator of joint $d$, then we can capture its evolution over time $\xi_d(t)= \sum_{d'=1}^{D} \sum_i a_{i,d'} k_{d,d'}(t,t_i)$, taking into account the effect of all other joints $d'$\@. 

The inner product in $\mathcal{H}$ of functions $\boldsymbol{\xi}^1(\cdot) =  \sum_{i,d} a_{i,d}k(t_i,\cdot)\bs{e}_d$ and $\boldsymbol{\xi}^2(\cdot) =  \sum_{j,d} b_{j,d}k(t_j,\cdot)\bs{e}_d$ is defined as: 
\meq{}{
&\langle \boldsymbol{\xi}^1,\boldsymbol{\xi}^2 \rangle_{\mathcal{H}}=\sum_d \langle \xi_d^1,\xi_d^2 \rangle_{\mathcal{H}}=\sum_d\sum_{i,j} a_{i,d}b_{j,d} k(t_i,t_j)\\ \label{eq:norm}
&\|\boldsymbol{\xi}\|^2_{\mathcal{H}} = \langle\boldsymbol\xi,\boldsymbol\xi\rangle = \sum_d \sum_{i,j} a_{i,d} a_{j,d} k(t_i,t_j)
}
For example, in the Gaussian RBF RKHS (with kernel $k_d(t,t') =\text{exp}(\|t-t'\|^2/2\sigma^2)$), a trajectory is a weighted sum of radial basis functions:
\meq{}{
\label{eq:rbftraj}
\boldsymbol{\xi}(t) =& \sum_{d,i} a_{i,d}\   \text{exp}\left(\frac{\|t-t_i\|^2}{2\sigma^2}\right)\bs{e}_d,\ \ {a_{i,d}}\in\mathbb{R}
}
The coefficients $a_{i,d}$ assess how important a particular joint $d$ at time $t_i$ is to the overall trajectory. They can be interpreted as weights of local perturbations to the motions of different joints centered at different times. The trajectory norm measures the size of the perturbations, and the correlation among them, quantifying how complex the trajectory is in the RKHS.

\section{Motion Planning in an RKHS}
\label{sec:trajopt}

In this section we describe how trajectory optimization can be achieved by functional gradient descent in an RKHS of trajectories.
\subsection{Cost Functional}
\label{sec:cost}
We introduce a cost functional $\mathcal{U}:\mathcal{H}\rightarrow \mathbb{R}$ that maps each trajectory to a scalar cost. This functional quantifies the quality of a given a trajectory (function in the RKHS). $\mathcal{U}$ trades off between a regularization term that measures the efficiency of the trajectory, and an obstacle term that measures its proximity to obstacles:
\meq{}{\label{eq:cost}
&\mathcal{U}[\boldsymbol{\xi}]= \mathcal{U}_{obs}[\boldsymbol{\xi}] +\frac{\beta}{2}\|\boldsymbol{\xi}\|_{\mathcal{H}}^2
 }
As described in \sref{sec:norm}, we choose our RKHS so that the regularization term encodes our desired notion of smoothness or trajectory efficiency (minimum length, velocity, acceleration, jerk).

The obstacle cost functional is defined on trajectories in configuration space, but obstacles are defined in the robot's workspace  $\mathcal{W}\equiv \mathbb{R}^3$.  So, we connect configuration space to workspace via a \emph{forward kinematics} map $x$: if $\mathcal B$ is the set of body points of the robot, then $x:\mathcal C\times\mathcal B\to\mathcal W$ tells us the workspace coordinates of a given body point when the robot is in a given configuration.  We can then decompose the obstacle cost functional as:
\meq{}{\label{eq:reduce}
{\mathcal{U}}_{obs}[\boldsymbol{\xi}] &= \mathop{\textrm{reduce}}_{t,u} c\left(x(\boldsymbol{\xi}(t),u)\right)
}
where $\textrm{reduce}$ is an operator that aggregates costs over the entire trajectory and robot body---e.g., a supremum or an integral, see \sref{sec:costa}. We assume that the $\textrm{reduce}$ operator takes (at least approximately) the form of a sum over some finite set of (time, body point) pairs $\boldsymbol{\mathcal T}(\boldsymbol\xi)$:
\meq{}{\label{eq:empirical}
{\mathcal{U}}_{obs}[\boldsymbol{\xi}] &= \sum_{(t,u)\in\boldsymbol{\mathcal T}(\boldsymbol\xi)} c\left(x(\boldsymbol{\xi}(t),u)\right)
}
For example, the supremum operator takes this form except on a measure-zero set of trajectories: whenever there is a unique supremum $(t,u)$, then $\boldsymbol{\mathcal T}(\boldsymbol\xi)$ is the singleton set $\{(t,u)\}$.  Integral operators do not take this form, but they can be well approximated in this form using quadrature rules, see \sref{sec:integralcost}.




\begin{algorithm}[t!]
   \caption{{\bf --- Trajectory optimization in RKHSs $\left(N, c,\nabla c,\boldsymbol{\xi}^{(n)}(0),\boldsymbol{\xi}^{(n)}(1)\right)$ }\label{alg:trajopt}}
\begin{algorithmic}[1]
\FOR {each joint angle $d\in D$} 
\STATE Initialize to a straight line trajectory ${\xi}^0_d(t)=\xi_d(0)+ (\xi_d(1)-\xi_d(0))t$.
\ENDFOR
\WHILE{($\mathcal{U}[\boldsymbol{\xi}^n]>\epsilon$  and  $n<N_{\rm MAX}$)}
	\STATE Compute $\mathcal{U}_{obs}[\boldsymbol{\xi}^n]$   \eref{eq:funccost}.
	\STATE Find the support $\boldsymbol{\mathcal{T}}(\bs{\xi})=\{t_i,u_i\},i=1,\dots,N$ time/body points \eref{eq:empirical}.
	\FOR {$(t_i,u_i)_{i=1}^{N}\in\boldsymbol{\mathcal T}(\boldsymbol\xi)$}
		\STATE Evaluate the gradient cost $ \nabla c(\boldsymbol{\xi}(t_i),u_i)$ and $\mathbf{J}(t_i,u_i)$ 
	\ENDFOR
	\STATE Update trajectory: \\ $\qquad\qquad\boldsymbol{\xi}^{n+1}= (1-\frac{1}{\lambda})\boldsymbol{\xi}^{n}-\frac{1}{\lambda}
\sum_{(t, u)\in\boldsymbol{\mathcal T}(\boldsymbol\xi)} \left(  \mathbf J^\top(t,u) \nabla c(x(\boldsymbol{\xi}(t),u))\right)^\top K(t,\cdot)$
	\STATE If constraints are present, project onto constraint set (\sref{eq:constsol}).	
\ENDWHILE
\STATE \textbf{\emph{Return:}} Final trajectory $\boldsymbol{\xi}^{*}$ and costs $\|\boldsymbol{\xi}\|^2_{\mathcal{H}}, \mathcal{U}_{obs}$.
\end{algorithmic}
\end{algorithm}

\subsection{Optimization}
\label{sec:opt}
We can derive the functional gradient update by minimizing a local quadratic approximation of $\mathcal U_{\rm obs}$:
\meq{}{\label{eq:mini}
\boldsymbol{\xi}^{n+1}=&\ \text{arg}\min\limits_{\boldsymbol{\xi}}\ \ \langle \boldsymbol{\xi}-\boldsymbol{\xi}^n, \nabla \mathcal{U}[\boldsymbol{\xi}^n] \rangle_{\mathcal{H}} + \frac{\lambda}{2}\|\boldsymbol{\xi}-\boldsymbol{\xi}^n\|^2_{\mathcal{H}}
}
The quadratic term is based on the RKHS norm, meaning that we prefer ``smooth'' updates, analogous to~\citet{Zucker13}
%
%
This minimization admits a solution in closed form:
\meq{}{\label{eq:update}
\boldsymbol{\xi}^{n+1}(\cdot)& = \left(1-\frac{1}{\lambda}\right)\boldsymbol{\xi}^n(\cdot) -\frac{1}{\lambda} \nabla \mathcal{U}_{obs}[\boldsymbol{\xi}^n](\cdot)
}
Since we have assumed that the cost functional $\mathcal{U}_{obs}[\boldsymbol{\xi}]$ depends only on a finite set of points $\boldsymbol{\mathcal{T}}(\boldsymbol\xi)$~\eref{eq:empirical},  it is straightforward to show that the functional gradient update has a finite representation (so that the overall trajectory, which is a sum of such updates, also has a finite representation).  In particular, assume the workspace cost field $c$ and the forward kinematics function $x$ are differentiable; then we can obtain the cost functional gradient by the chain rule:~\cite{Ratliff07,smolabook}
\meq{}{\label{eq:funccost}
\nabla\mathcal{U}_{obs}(\cdot)=\sum_{(t, u)\in\boldsymbol{\mathcal T}(\boldsymbol\xi)}
  \left(\mathbf J^\top(t,u) \nabla c(x(\boldsymbol{\xi}(t),u))\right)^{\top}K(t,\cdot)
}
where $\mathbf J(t,u) = \frac{\partial}{\partial \boldsymbol\xi(t)}x(\boldsymbol\xi(t),u) \in \mathbb{R}^{3\times D}$ is the workspace Jacobian matrix at time $t$ for body point $u$, so that the kernel function $K(t,\cdot)$ is the gradient of $\boldsymbol\xi(t)$ with respect to $\boldsymbol\xi$. The kernel matrix is fully defined in Equation \eref{eq:vecrkhs}.

This solution is a generic form of functional gradient optimization with a \emph{directly instantiable} obstacle gradient that does not depend on a predetermined set of waypoints, offering a more expressive representation with fewer parameters. We derive a constrained optimization update rule, by solving the KKT conditions for a vector of Lagrange multipliers, see \sref{sec:constraints}.  The full method is summarized as \aref{alg:trajopt}. 



\section{Trajectory Efficiency as Norm Encoding in RKHS}
\label{sec:norm}
In different applications it is useful to consider different notions of trajectory efficiency or smoothness.  We can do so by choosing RKHSs with appropriate norms.  
For example, it is often desirable to penalize the velocity, acceleration, jerk, or other derivatives of a trajectory instead of (or in addition to) its magnitude.  To do so, we can take advantage of a \emph{derivative reproducing property}: let $\mathcal H_1$ be one of the coordinate RKHSs from our trajectory representation, with kernel $k$. If $k$ has sufficiently many continuous derivatives, then for each partial derivative operator $D^\alpha$, there exist representers $(D^\alpha k)_t\in\mathcal H_1$ such that, for all $f\in\mathcal H_1$,
$(D^\alpha f)(t) = \langle (D^\alpha k)_t, f\rangle$~\cite[Theorem 1]{zhou08}.  (Here $\alpha$ is a multi-set of indices, indicating which partial derivative we are referring to.)
We can therefore define a new RKHS with a norm that penalizes the partial derivative $D^\alpha$: the kernel is $k^\alpha(t, t') = \langle (D^\alpha k)_t, (D^\alpha k)_{t'}\rangle$.  
If we use this RKHS norm as the smoothness penalty for our trajectories, then our optimizer will automatically seek out trajectories with low velocity, acceleration, or jerk.

Consider an RBF kernel with a reproducing first order derivative: $D^{1}k(t,t_i) =D^1k_{t_i}[t]= \frac{(t-t_i)}{2 \sigma^2} k(t,t_i)$ is the reproducing kernel for the velocity profile of a trajectory defined in an RBF kernel space $k(t,t_i)= \frac{1}{\sqrt{2\pi\sigma^2}}\exp(-\|t-t_i\|^2/2\sigma^2)$.
The velocity profile can be written as $D^{1}\boldsymbol{\xi} (t) = \sum_i \beta_i D^{1} k(t,t_i)$, with endpoint conditions $D^1 \boldsymbol{\xi}(0)=\dot{\boldsymbol{q}}_i,\ D^1\boldsymbol{\xi}(1) = \dot{\boldsymbol{q}}_f$.

The trajectory can be found by integrating $D^1\boldsymbol{\xi}(t)$ once and projecting onto the nullspace of the constraints $\boldsymbol{\xi}(0)=\boldsymbol{q}_i,\boldsymbol{\xi}(1)=\boldsymbol{q}_f$.
\meq{}{
\boldsymbol{\xi}(T) &= \int\limits_0^1 D^{1} \boldsymbol{\xi}(t) dt = \sum_i \beta_i \int\limits_0^1  \frac{(t-t_i)}{2 \sigma^2}  k(t,t_i) dt= \sum_i \beta_i \left[k(T,t_i) - k(0,t_i)\right] +\boldsymbol{q}_i
}
The initial condition is verified automatically and the endpoint condition can be written as $\boldsymbol{q}_f = \sum_i \beta_i \left[k(1,t_i) - k(0,t_i)\right] +\boldsymbol{q}_i$, this imposes additional information over the coefficients $\beta_i\in \mathcal{C}$.
Here we explicitly considered only a $\mathcal{H}^1$ space, but extensions to higher order derivatives can be derived similarly integrating p times to obtain the trajectory profile. Constraints over higher derivatives can be computed using any constraint projection method. The update rule in this setting can be derived using the natural gradient in the space, where the new obstacle gradient becomes:
\meq{}{
\nabla U_{obs}[\boldsymbol{\xi}](t) &= \sum_j^n\sum_{(t_i,u_i)\in \mathcal{T}} \left( \mathbf{J}^\top(t_i,u_i) \nabla c(\ x(\boldsymbol{\xi}(t_i),u_i)\ )\right)^\top D^{j} k(t_i,t)\ \ \label{eq:dergrad}
}
Regularization schemes in different RKHSs may encode different forms of trajectory efficiency. We provide a form of penalizing trajectory complexity in different forms by minimizing the trajectory norm in the RKHS. This may be defined in terms of the reproducing kernel, by sums, products, tensor product of kernels, or any closed kernel operation.

 \subsection{Kernel Metric in RKHS}
 \label{sec:vecnorm}
 The norm provides a form of quantifying how complex a trajectory is in the space associated with the RKHS kernel metric $K$. The kernel metric is determined by the kernel functions we choose for the RKHS, as we have seen before (\sref{sec:norm}). Likewise, the set of time points $\mathcal{T}$ that support the trajectory contribute to the design of the kernel metric:
  \meq{}{
 \|\bs{\xi}\|_{\mathcal{H}}^2&=\sum_{d}\sum_{t_i,t_j\in \mathcal{T}} {a}_{d,i} k_{d,d'}(t_i,t_j) {a}_{d',j}\nt
 &=\sum_{t_i,t_j\in \mathcal{T}} \bs{a}_{i}^\top K(t_i,t_j) \bs{a}_{j'},\ \ \bs{a}_i,\bs{a}_j\in\mathbb{R}^{D}\nt
 &=\bs{a}^\top \bs{K}(\mathcal{T},\mathcal{T})\bs{a}, \bs{a}\in\mathbb{R}^{DN}
 }
 Here $\bs{a}$ is the concatenation of all coefficients $\bs{a}_i$ over $\mathcal{T},\ |\mathcal{T}|=N$.  $\bs{K}(\mathcal{T},\mathcal{T})\in\mathbb{R}^{DN\times DN}$ is the \emph{Gram matrix} for all time points in the support of $\bs{\xi}$, and all joint angles of the robot. This matrix expresses the degree of correlation or similarity among different joints throughout the time points in $\mathcal{T}$.
 It can be interpreted, alternatively, as a tensor metric in a Riemannian manifold.~\cite{Amari98,Ratliff15}  Its inverse is the key element that bridges the gradient of functional cost $\nabla \mathcal{U}$ (gradient in the RKHS, Eq.\ref{eq:funccost} ), and its conventional gradient (Euclidean gradient).\footnote{This is what makes the optimization process covariant (invariant to reparametrization).}
\meq{}{
\nabla \mathcal{U} = \bs{K}^{-1}(\mathcal{T},\mathcal{T}) \nabla_{E}\ \mathcal{U}\nt
}
 The minimizer of the full functional cost $\mathcal{U}$ has a closed form solution in \eref{eq:update}. Where the gradient $\nabla \mathcal{U}$, is the natural gradient in the RKHS. This can be seen as a warped version of the obstacle cost gradient according to the RKHS metric.

\section{Cost Functional Analysis}
\label{sec:costa}
Next we analyze how the cost functional (different forms of the reduce operation in \sref{sec:cost}), affect obstacle avoidance performance, and the resulting trajectory (\sref{sec:costa}). 
In this paper, we adopt a maximum cost version (\sref{sec:maxcost}), and an approximate integral cost version of the obstacle cost functional (\sref{sec:integralcost}). Other variants could be considered, providing the trajectory support remains finite, but we leave this as future work. Additionally, we compare the two forms (\sref{sec:integralcost-experiments}), against a more commonly used cost functional, the path integral cost,~\cite{Ratliff09} and we show our formulations do not perform worse, while being faster to compute.
Based on these experiments, in the remaining sections of the paper we consider only the max cost formulation, which we believe represents a good tradeoff between speed and performance. 

\subsubsection{Max Cost Formulation}
\label{sec:maxcost}
\begin{figure}[ht!]
\centering
\includegraphics[width=.6\textwidth]{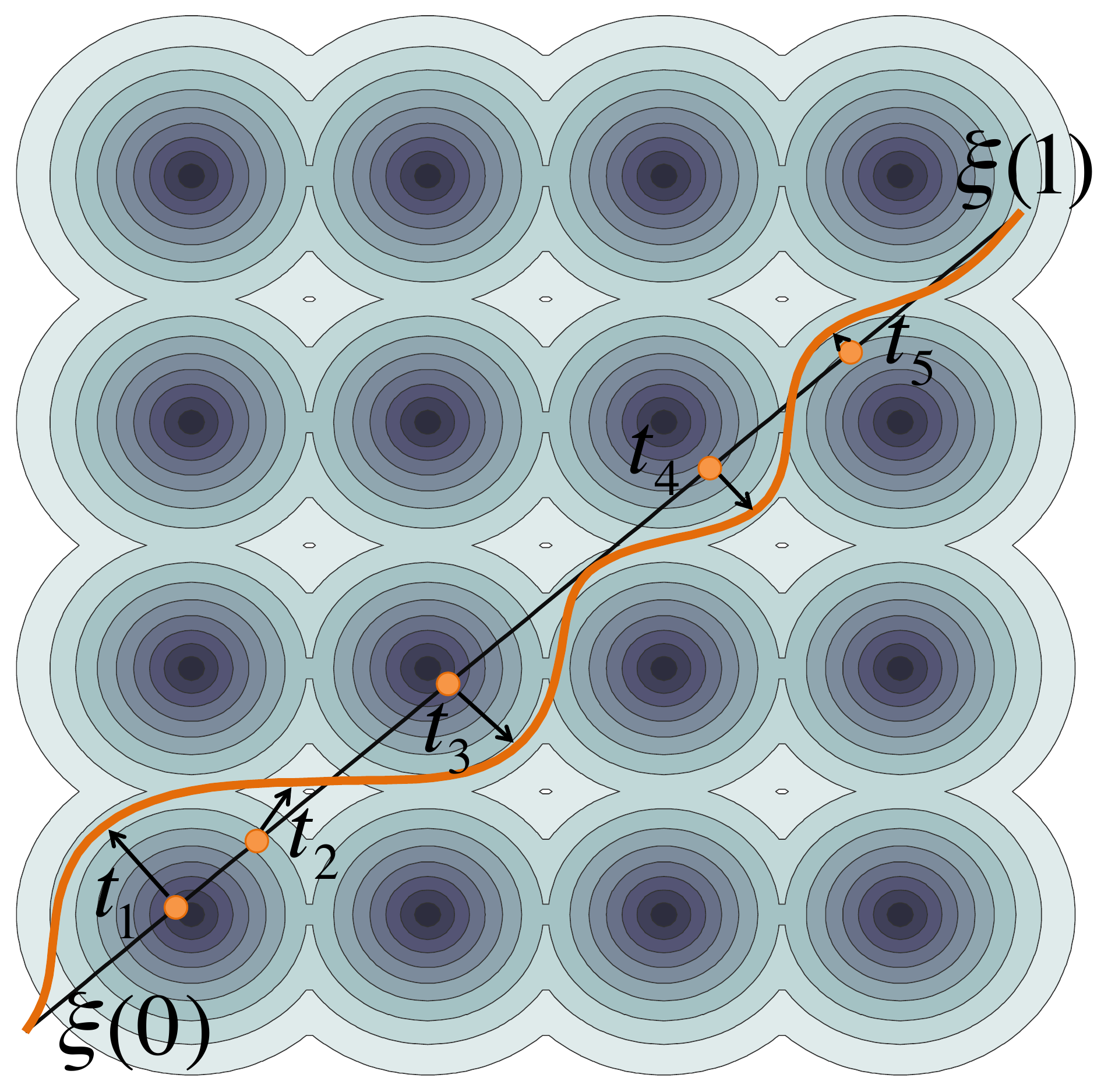}
\caption{\footnotesize At every iteration, the optimizer takes the current trajectory (black) and identifies the point of maximum obstacle cost $t_i$ (orange points). It then updates the trajectory by a point evaluation function centered around $t_i$. Grey regions depict isocontours of the obstacle cost field (darker means closest to obstacles, higher cost).}
\label{fig:1}
\end{figure}

The maximum obstacle cost penalizes points in the trajectory close to obstacles, \emph{i.e.} high cost regions in workspace (regions inside/near obstacles). This maximum cost version of the reduce operation, considered in Eq.\eref{eq:reduce}, can be described as picking time points (sampling), deepest inside or closest to obstacles, see \figref{fig:1}. 
The sampling strategy for picking time points to represent the trajectory cost can be chosen arbitrarily, and further improved for time efficiency. In this paper, we consider a simple version, where we sample points along sections of the trajectory, and choose $Nx$ maximum violating points, one per section.

This max cost strategy allows us to represent trajectories in terms of a few points, rather then a set of finely discretized waypoints. This is a simplified version of the obstacle cost functional, that yields a more compact representation.\cite{Ratliff09,park2012itomp,stomp} 

\subsubsection{Integral Cost Formulation}
\label{sec:integralcost}
Instead of scoring a trajectory by the supremum of obstacle cost over time and body points, it is common to integrate cost over the entire trajectory and body, with the trajectory integral weighted by arc length to avoid velocity dependence.~\cite{Zucker13}  While this path integral depends on all time and body points, we can approximate it to high accuracy from a finite number of point evaluations using numerical quadrature.~\cite{numericalrecipes}  $\boldsymbol{\mathcal{T}}(\boldsymbol\xi)$ then becomes the set of abscissas of the quadrature method, which can be adaptively chosen on each time step (e.g., to bracket the top few local optima of obstacle cost), see \sref{sec:pathintegralapprox}.  In our experiments, we have observed good results with Gauss-Legendre quadrature.

\subsection{Integral vs. Max cost Formulation}
\label{sec:integralcost-experiments}
\begin{figure*}[ht!]
\centering
\subfigure[$\mathcal{U}_{obs}$, Integral vs Max cost]{
\includegraphics[width=.4\textwidth]{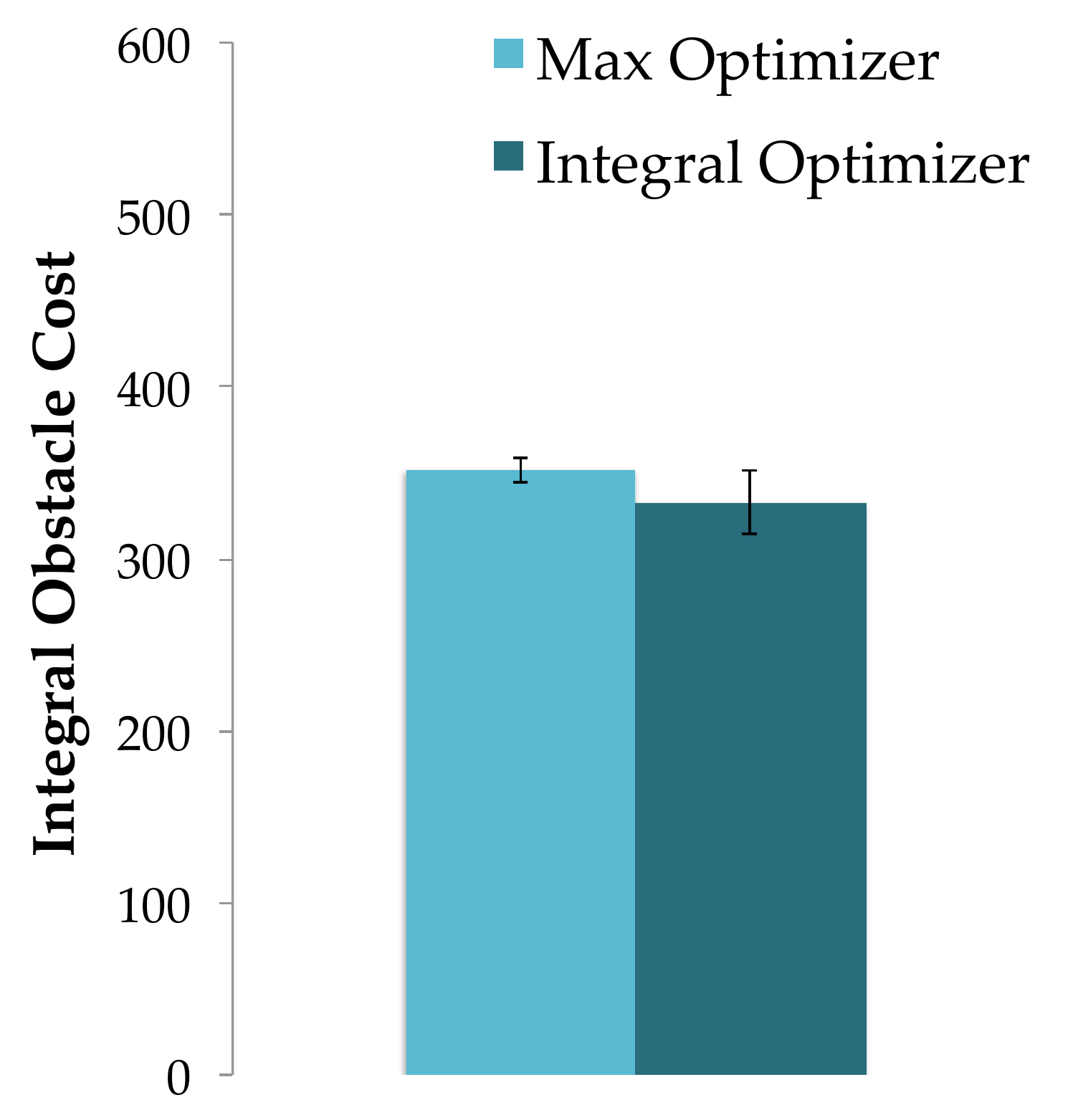}\label{fig:maxint}}
\subfigure[$\mathcal{U}_{obs}$, Approx integral vs Max cost]{
\includegraphics[width=.4\textwidth]{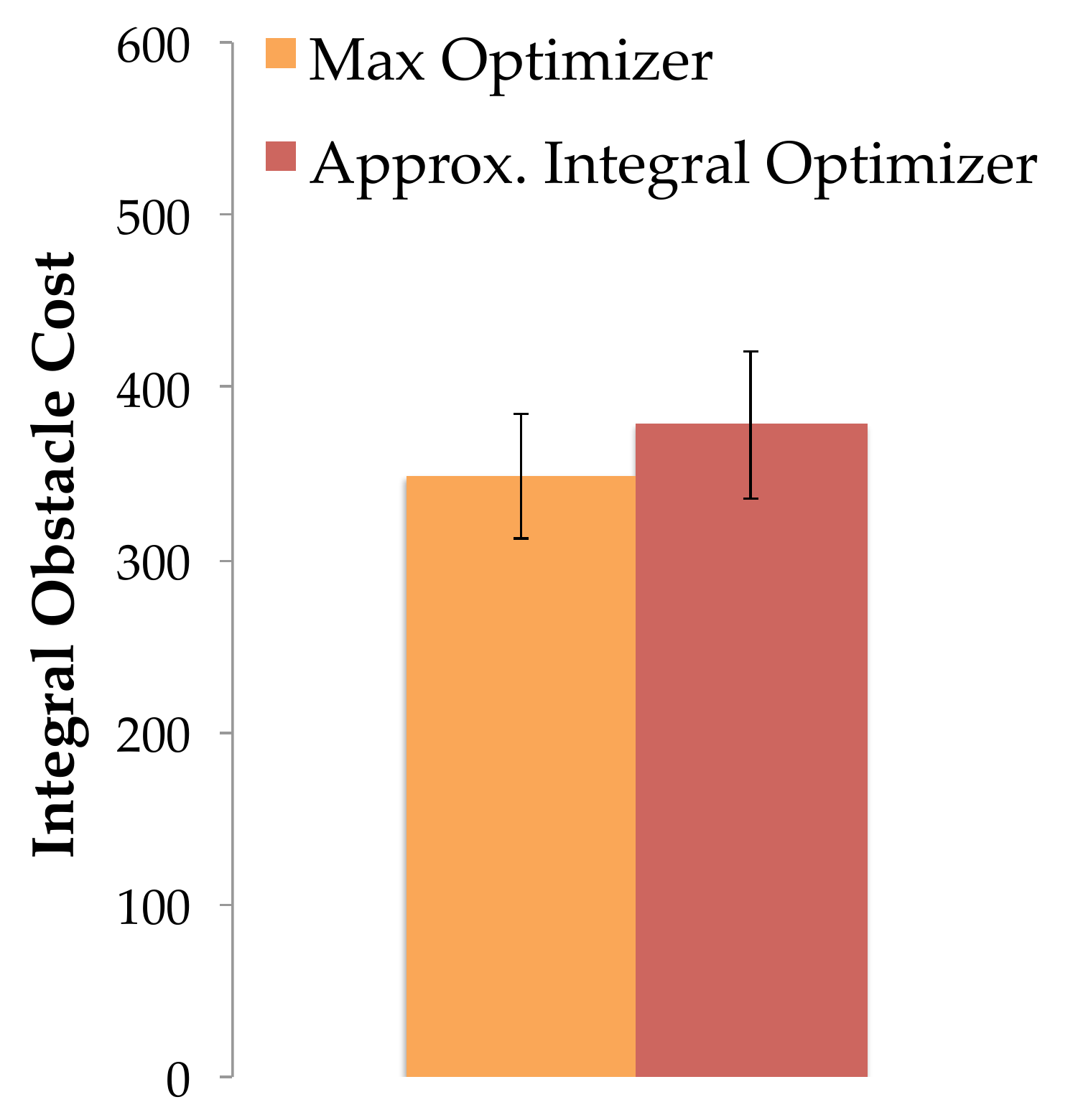}\label{fig:maxaint}}
\caption{ a) The integral costs after 5 large steps comparing between optimizing using our obstacle cost formulation with Gaussian RBG kernels vs. the integral formulation (using waypoints). b) A comparison between Gaussian RBF kernel integral cost using our max formulation vs. the approximate quadrature cost (20 points, 10 iterations).} 
\label{fig:comparisons}
\end{figure*}
We show that our new formulation does not hinder the optimization -- that it leads to practically equivalent results as an integral over time and body points.~\cite{Zucker13} To do so, we manipulate the cost functional formulation, and measure the resulting trajectories' cost in terms of the integral formulation. \figref{fig:maxint} shows the comparison: the integral cost increased by only $5\%$ when optimizing for the max.
Additionally we tested the max cost formulation against the approximate integral cost using a Gauss-Legendre quadrature method. We performed tests over 100 randomly sampled scenarios and measured the final obstacle cost after 10 iterations. We used 20 points to represent the trajectory in both cases. \figref{fig:maxaint} shows the approximate integral cost formulation is only $8\%$ above the max approach.

%
%

\section{Experimental Results}
\label{sec:results}
In what follows, we compare the performance of RKHS trajectory optimization vs. a discretized version (CHOMP) on a set of motion planning problems in a 2D world for a 3 DOF link planar arm as in \figref{fig:kernels}, and how different kernels with different norms affect the performance of the algorithm~(\sref{sec:main_experiment}).
We then, introduce a series of experiments that illustrate why RKHSs improve optimization (\sref{sec:larger_steps}). 

\subsection{RKHS with Radial Basis  vs. Waypoints}
\label{sec:main_experiment}
For our main experiment, we systematically evaluate the two parameterizations across a series of planning problems.
Although, Gaussian RBFs are a default choice of kernel in many kernel methods, RKHSs can also easily represent other types of kernel functions, \emph{e.g.} For example, B-splines are a popular parameterization of smooth functions,~\cite{ZK95,PZM95,Blake98} that are able to express smooth trajectories while avoiding obstacles, even though they are finite dimensional kernels. 
The choice of kernel should be application driven, and any reproducing kernel can easily be considered under the the optimization framework presented in this paper.

In the following experiment, we manipulate the parameterization (waypoints vs different kernels) as well as the number of iterations (which we use as a covariate in the analysis).  To control for the cost functional as a confound, we use the max formulation for both parameterizations. We use iterations as a factor because they are a natural unit in optimization, and because the amount of time per iteration is similar: the computational bottleneck is computing the maximum penetration points. 
We measure the obstacle and smoothness cost of the resulting trajectories. For the smoothness cost, we use the norm in the waypoint parameterization as opposed to the norm in the RKHS as the common metric.\\
\begin{figure*}[ht!]
\centering
\subfigure[Obstacle cost vs. kernel choice]{
\includegraphics[width=.48\textwidth]{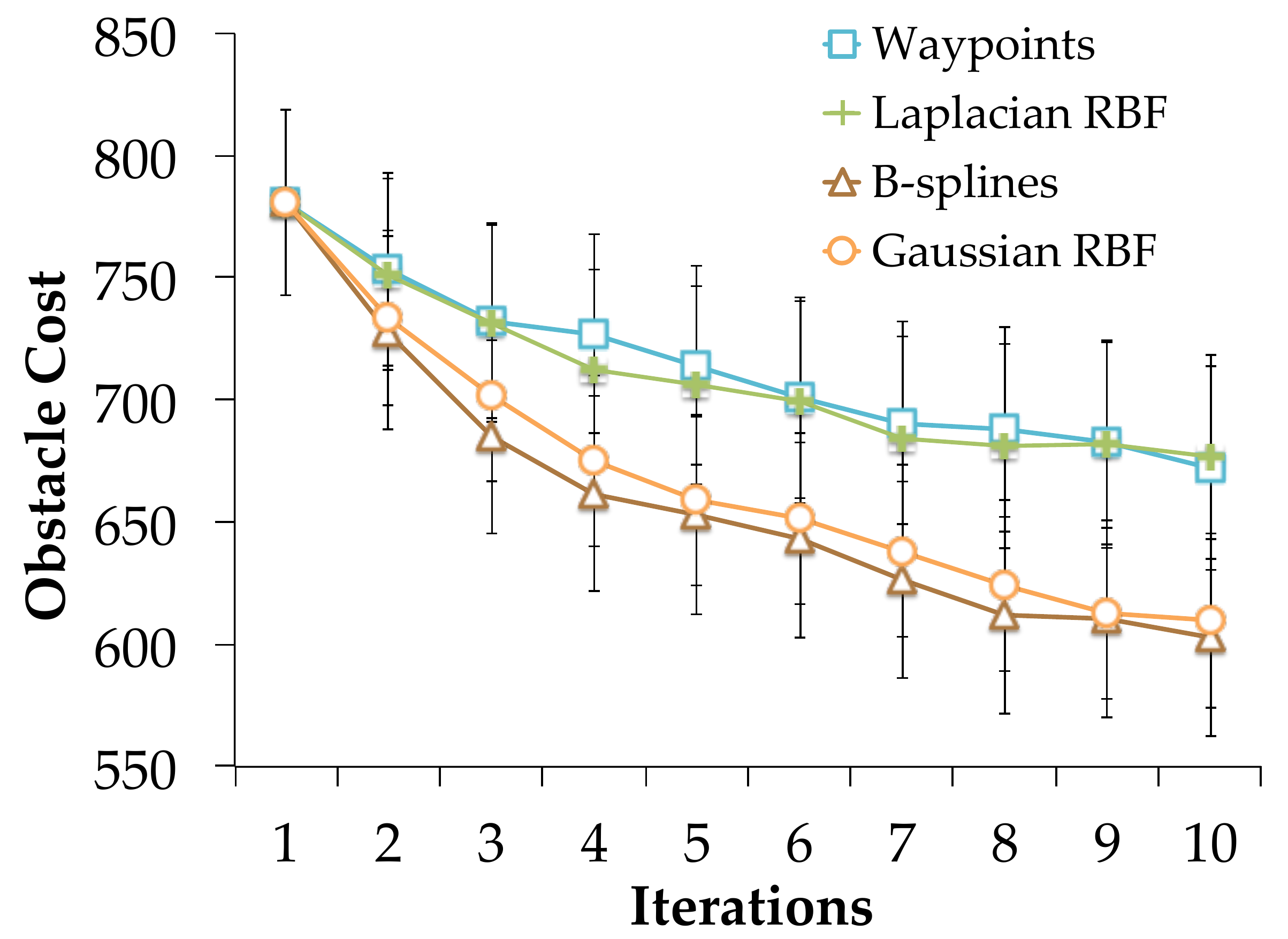}}
\subfigure[Smoothness cost vs. kernel choice]{
\includegraphics[width=.48\textwidth]{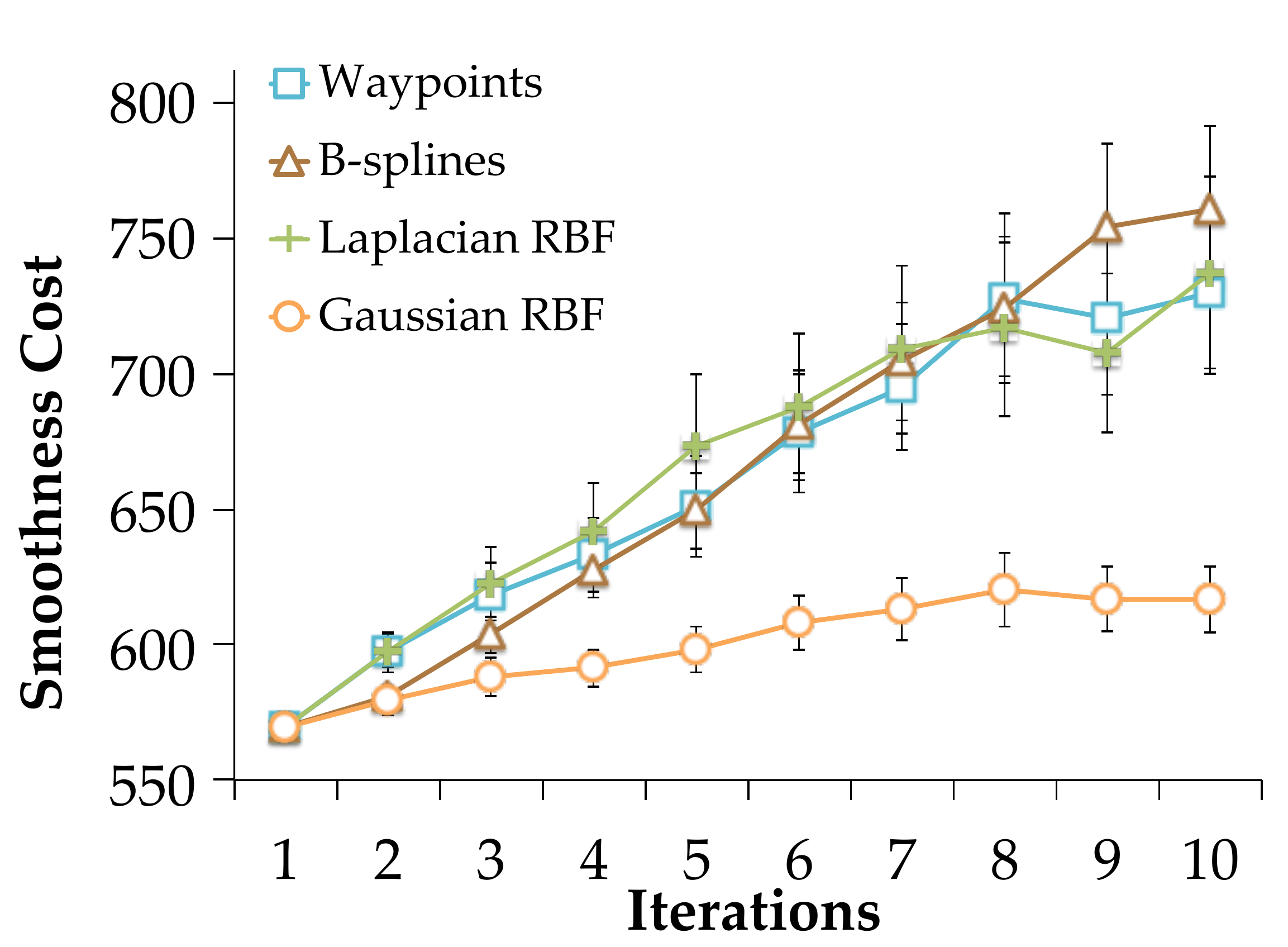}}
\caption{ Cost over iterations for a 3DoF robot in 2D. Error bars show the standard error over 100 samples. }
\label{fig:kernelcost}
\end{figure*}
\begin{figure*}[ht!]
\centering
\subfigure{
\includegraphics[width=.9\textwidth]{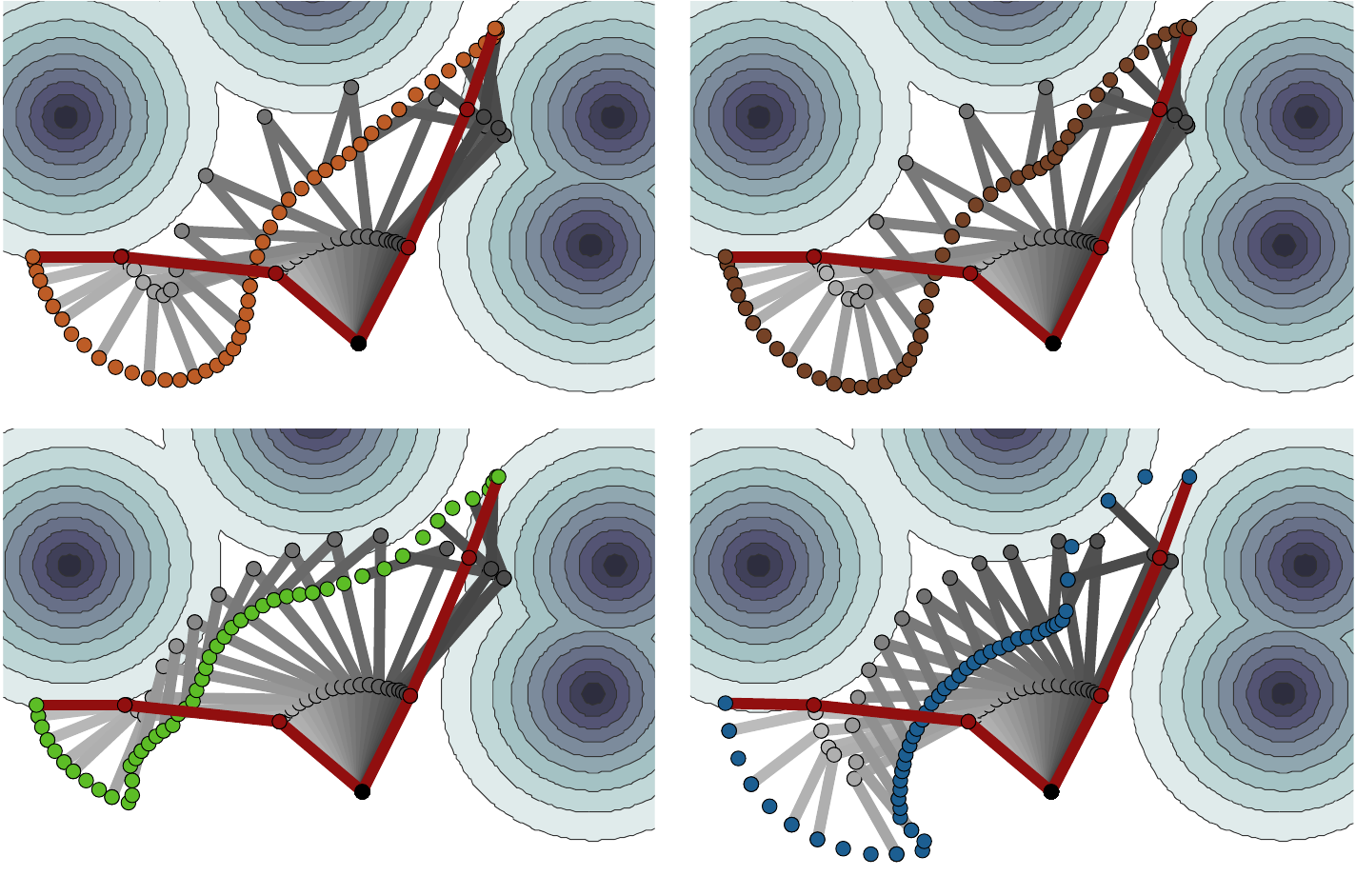}}
\caption{Robot 3DoF in C-space. Trajectory after 10 iterations: top-left: Gaussian RBF kernel, top-right: B-splines kernel, bottom-left: Laplaceian RBF kernel, bottom-right: Waypoints.} 
\label{fig:kernels}
\end{figure*}

The RKHS parameterization results in comparable obstacle cost and lower smoothness cost for the same number of iterations.
 We use 100 different random obstacle placements and keep the start and goal configurations fixed as our experimental setup. The trajectory is represented with 4 maximum violation points over time and robot body points. In the analysis we performed a t-test using the last iteration samples, and showed that the Gaussian RBF RKHS representation resulted in significantly lower obstacle cost ($t(99)=-2.63$, $p<.01$) and smoothness cost ($t(99)=-3.53$, $p<.001$), supporting our hypothesis. We expect this to be true because with the Gaussian RBF parameterization can take larger steps without breaking smoothness, see \sref{sec:larger_steps}.
 
 We observe that Waypoints and Laplacian RBF kernels with large widths have similar behavior, while Gaussian RBF and B-splines kernels provide a smooth parameterization that allows the algorithm to take larger steps at each iteration. These kernels provide the additional benefit of controlling the motion amplitude, being the most suitable in the implementation of an adaptive motion planner. 
Laplacian RBF kernels yield similar results as the waypoint parameterization, since it is less affected by the choice of the width of the kernel. \figref{fig:kernels} provides a qualitative evaluation of the effect of different kernel choices. We compare the effectiveness of obstacle avoidance over 10 iterations, in 100 trials, of 12 randomly placed obstacles in a 2D environment, see \figref{fig:kernels}.

\subsection{RKHSs Allow Larger Steps than Waypoints}
\label{sec:larger_steps}
\label{fig:largesteps}

\begin{figure*}[ht!]
\centering
\subfigure[top: Gaussian RBF large steps (5 it.); middle: waypoints large steps (5 it.); bottom: waypoints small steps (25 it.)]{
\includegraphics[width=.52\textwidth]{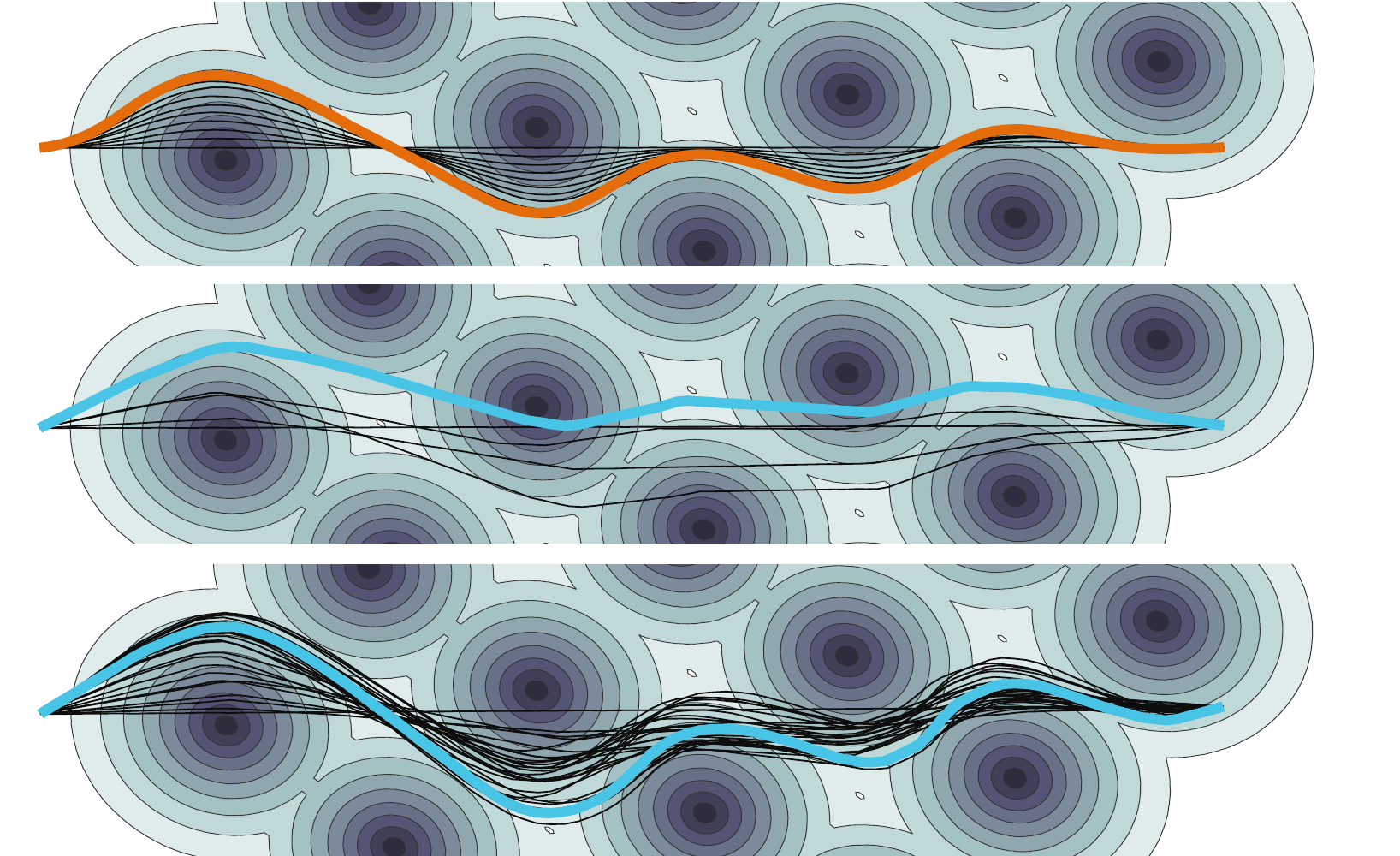}
\label{fig:comparisons}}
\subfigure[in order: Gaussian RBF, B-splines, Laplacian RBF kernels, and waypoints with large steps for 1 iteration.]{
\includegraphics[width=.44\textwidth]{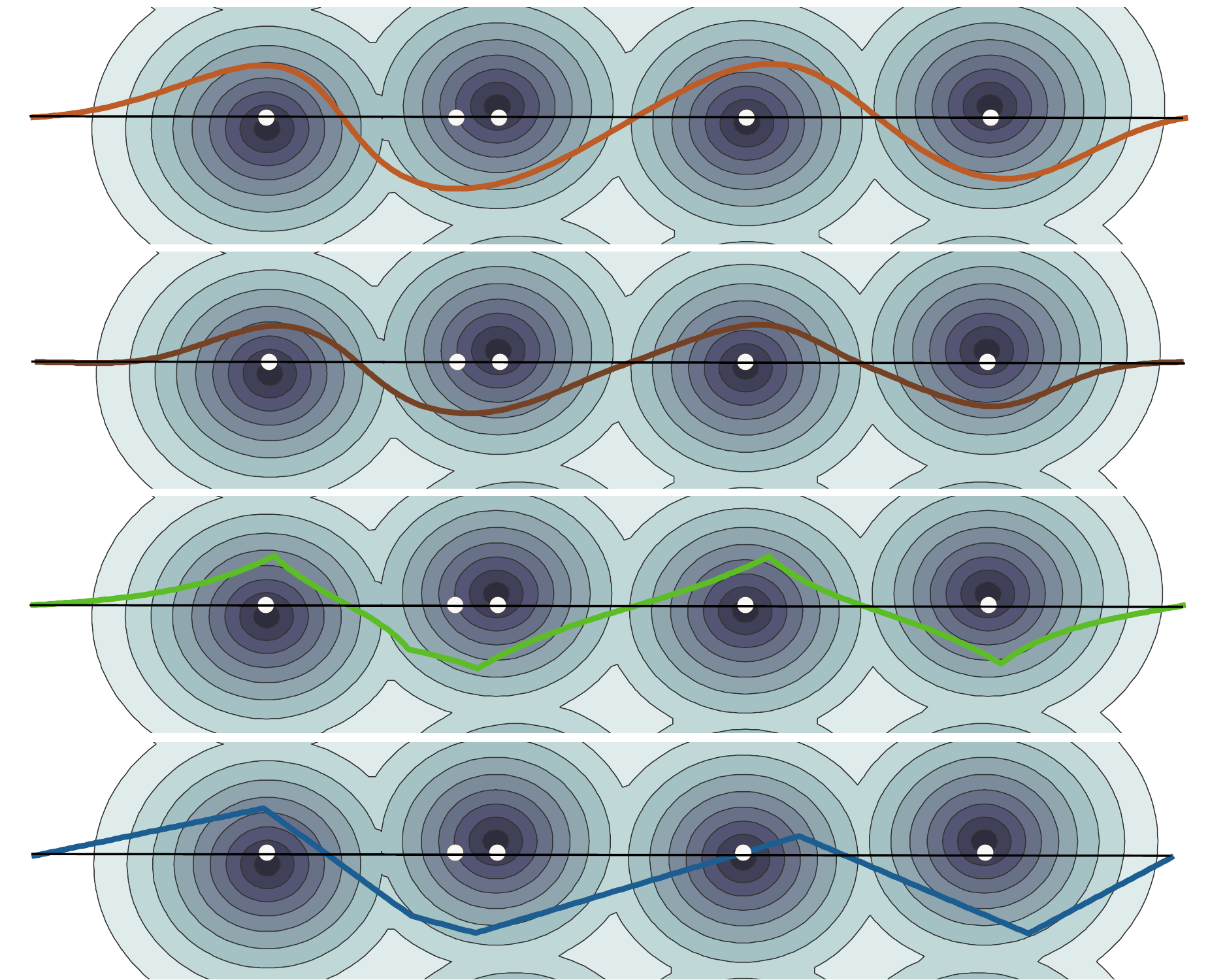}
\label{fig:largesteps}}
\caption{ a) 2d trajectory of 1dof robot in a maze environment (obstacle in shaded grey). b)Trajectory profile using different kernels (5 time points in white). } 
\end{figure*}
One practical advantage of using an Gaussian RBF RKHS instead of the waypoint parameterization is the ability to take large steps during the optimization. \figref{fig:comparisons} compares the two, while taking large steps: it takes 5 Gaussian RBF iterations to solve the problem, but would take 28 iterations with smaller steps for the waypoint parameterization -- otherwise, large steps cause oscillation and break smoothness. The resulting obstacle cost is always lower with Gaussian RBFs ($t(99)=5.32$, $p<.0001$). The smoothness cost is higher ($t(99)=8.86$, $p=<.0001$), as we saw in the previous experiment as well-- qualitatively,  however, as \figref{fig:largesteps} shows, the Gaussian RBF trajectories appear smoother, even after just one iteration, as they do not break differential continuity. 
So far, we used 100 waypoints to represent the trajectory, and only 5 kernel evaluation points for the RKHS. We did also test the waypoint parameterization with 5 waypoints, in order to have an equivalently low dimensional representation. This resulted in much poorer behavior with regards to differential continuity.

%
\subsection{Real World Experiments on a 7-DOF Manipulator}
\begin{figure*}[ht!]
\centering
\subfigure[Gaussian RBF (orange) vs. Waypoints (blue)]{
\includegraphics[width=.48\textwidth]{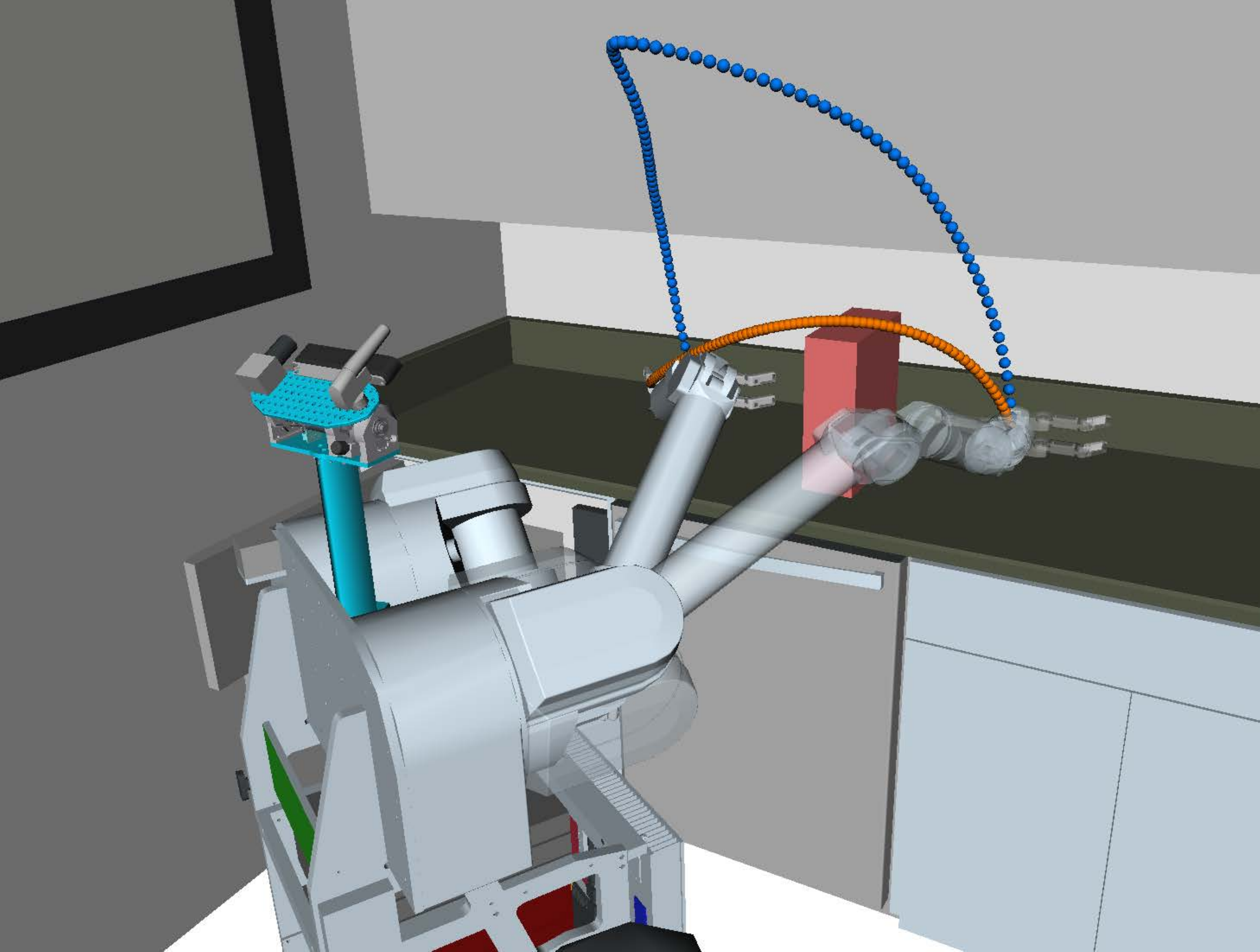}\label{fig:7dof2}}
\subfigure[Gaussian RBF (orange) vs. Waypoints (blue)]{
\includegraphics[width=.445\textwidth]{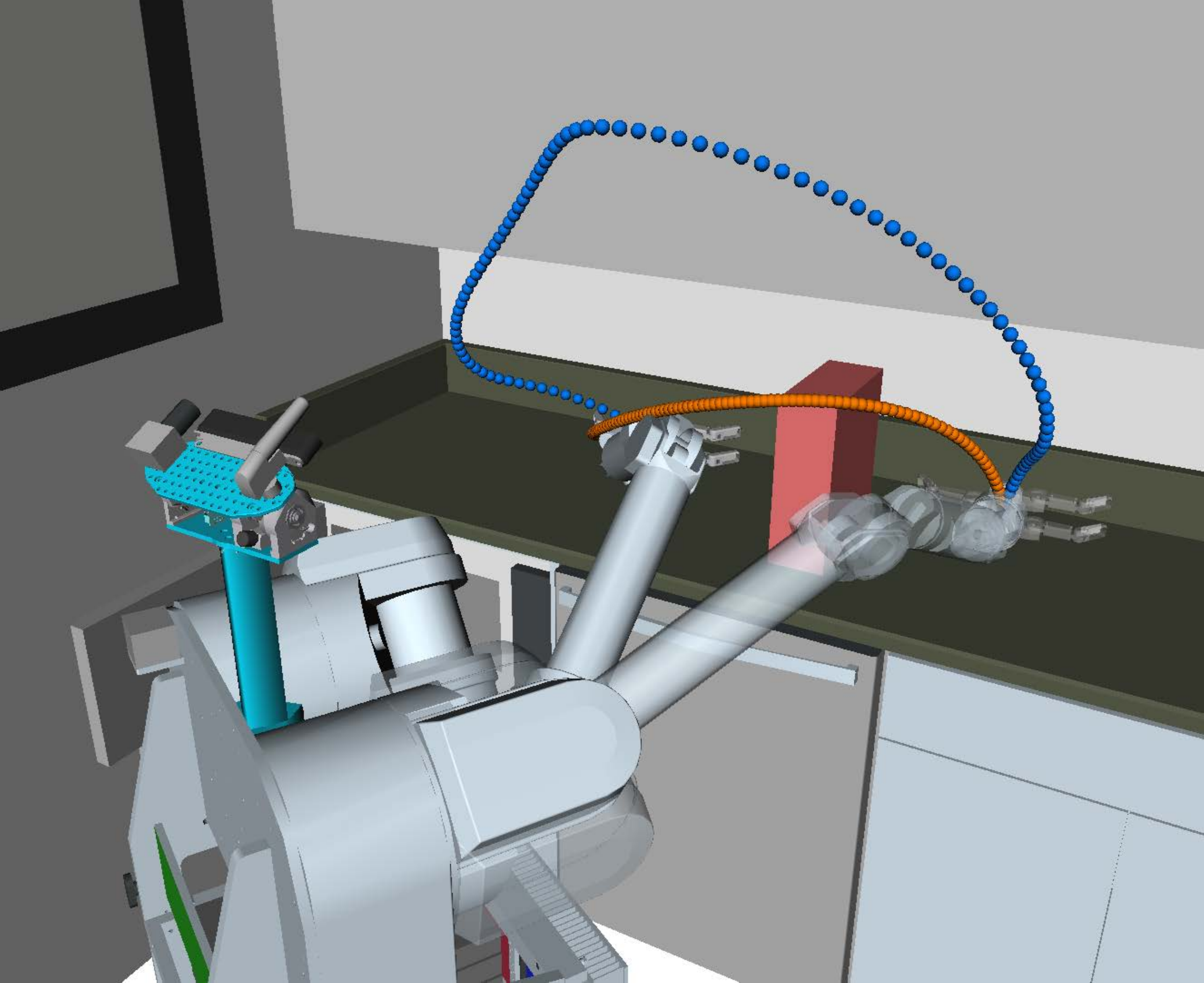}\label{fig:7dof}}
\caption{ 7-dof robot experiment, plotting end-effector position from start to goal. (a) Gaussian RBF RKHS with 1 max point (10 iterations, $\lambda=20,\beta=0.5$) vs. Waypoints (10 iterations, $\lambda=200$). (b) Gaussian RBF RKHS with 1 max point (25 iterations, $\lambda=20,\beta=0.5$) vs. Waypoints (25 iterations, $\lambda=200$).
\label{fig:7dofsd}} 
\end{figure*}

\figref{fig:7dofsd} shows a comparison between the waypoint parametrization (CHOMP) and the RKHS Gaussian RBF on a 7-DOF manipulation task. \figref{fig:7dof2} shows the end-effector traces, after 10 iterations of optimization, for both methods. The path for CHOMP (blue) is very non-smooth and collides with the counter while the Gaussian RBF optimization is able to find a smoother path (orange) that is not in collision. Note that we only use a single max-point for the RKHS version, which leads to much less computation per iteration, as compared to CHOMP. \figref{fig:7dof} shows the results from both methods after 25 iterations of optimization. CHOMP is now able to find a collision-free path, but the path is still not very smooth as compared to the RKHS-optimized path. These results echo our findings from the robot simulation and planar arm experiments. 
We are currently looking at more experiments in these high-dimensional configuration spaces, where we believe the RKHS approach with its better representative power can find smoother collision-free paths faster.

%
%
\section{Discussion and Future Work}
\label{sec:discussion}
 In this work we presented an expressive kernel approach to trajectory representation: we represent smooth trajectories as vectors in an RKHS\@.  Different kernels lead to different notions of smoothness, including commonly-used variants as special cases.
We introduced a functional gradient trajectory optimization method based on our RKHS representation, and demonstrated that this optimizer can take large steps, leading to a smooth and collision-free trajectory faster than optimizers that use less-flexible representations.
We can think of the functional gradient iteration as automatically adapting the temporal resolution of our trajectory during optimization.

Our work is only the first step in exploring RKHSs for motion planning. In the future, we are excited about the potential of this work for both learning from experience and
learning from demonstration.
First, a low-dimensional trajectory parameterization enables us to more easily generate a diverse set of initial trajectories for an optimizer, aiding techniques that learn how to score initial trajectories for a new motion planning problem based on data from old problems.~\cite{Dey12}
Second, RKHSs enable us to plan with kernels learned from user demonstrations, leading to spaces in which more predictable motions have lower norm, and ultimately fostering better human-robot interaction.~\cite{dragan14fam}


\clearpage
{\small \bibliographystyle{plainnat}
\bibliography{references}}
 
 \clearpage
\appendix
\section{Appendix}
\label{sec:app}

\subsection{Finite approximation of Path Integral Cost}
\label{sec:pathintegralapprox}
Trajectory optimization in RKHSs can be derived for different types of obstacle cost functionals, provided that trajectories have a finite representation.
 Previous work defines a obstacle cost in terms of the arc-length integral of the trajectory.~\cite{Zucker13} We approximate the path integral cost functional, with a finite representation using integral approximation methods, such as quadrature methods.~\cite{numericalrecipes}Consider a set of finite time points $t_i\in \boldsymbol{\mathcal{T}}$ to be the abscissas of an integral approximation method. We use a Gauss-Legendre quadrature method, and represent $t_i$ as roots of the Legendre polynomial $P_n(t)$ of degree $n$. Let $w_i$ be the respective weights on each cost sample:
 \meq{}{\label{eq:funcintcost}
 \mathcal{U}_{obs}[\boldsymbol{\xi}]  = \int\limits_0^1  c\left[\boldsymbol{\xi}(t) \right] \left\|D^1\boldsymbol{\xi}(t)\right\| dt &\approx \sum_{t_i\in\mathcal{T}}  \omega_i\ c\left[\boldsymbol{\xi}(t_i) \right] \left\|D^1\boldsymbol{\xi}(t_i)\right\|
 }
 with coefficients, and the Legendre polynomials obtained recursively from the Rodriguez Formula:
\begin{align*}
 P_n &=  2^{n}\sum\limits_{j=0}^{n} t^{j} \dbinom{n}{j} \dbinom{\frac{n+j-1}{2}}{n} \nt
  w_i &= \frac{2}{\left( 1-t_i^2 \right) [D^1P_n(t_i)]^2}
 \end{align*}
 We denote $D^1\equiv \frac{d}{dt}$ the first order time derivative. Using this notation, we are able to work with integral functionals, using still a finite set of time points to represent the full trajectory.

 \subsection{Waypoint Parameterization as an Instance of RKHS}
 \label{sec:waypoints}
Consider a general Hilbert space of trajectories $\xi \in \Xi$, (not necessarily an RKHS) equipped with an inner product $\langle \xi_1,\xi_2\rangle_{\Xi} =  \xi_1^T A \xi_2 $. In the waypoint representation,~\cite{Zucker13} $A$ is typically the Hessian matrix over points in the trajectory, which makes the norm in $\xi$ penalize unsmooth and inefficient trajectories, in the sense of high acceleration trajectories.
 The minimization under this norm $\|\xi\|_A=\sqrt{\xi^TA\xi}$ performs a line search over the negative gradient direction, where $A$ dictates the shape of the manifold over trajectories.
  This paper generalizes the waypoint parameterization, we can represent waypoints by representing the trajectory in terms of delta Dirac basis functions $\langle \xi, \delta(t,\cdot)\rangle= \xi(t)$ with an additional smoothness metric $A$. Without A, each individual point is allowed to change without affecting points in the vicinity. Previous work, overcome this caveat by introducing a new metric that propagates changes of a single point in the trajectory to all the other points. Kernel evaluations in this case become $k(t_i,\cdot)=A^{-1}\delta(t_i,\cdot)$, where $\xi(t) = \sum_i a_i A^{-1}\delta(t_i,\cdot)$. The inner product of two functions is defined as $\langle \xi_1,\xi_2\rangle_{A} = \sum_{i,j}  a_ib_i A^{-1}\delta(t_i,t_i) $.\\
Here $\delta(t_i,\cdot)$ represents the finite dimensional delta function which is one for point $t_i$ and zero for all the other points. A trajectory in the waypoint representation becomes a linear combination of the columns of $A^{-1}$. Columns of $A^{-1}$ dictate how the corresponding point will affect the full trajectory.\\
For an arbitrary kernel representation the behavior of these points over the full trajectory are associated with the kernel functions associated with the space. For radial basis functions the trajectory is represented as gaussian functions centered at a set of chosen time points (fewer in practice) instead of the full trajectory waypoints.
In this sense, we have a more compact trajectory representation using RKHSs.

 \subsection{Constrained optimization}
 \label{sec:constraints}
 Consider equality and inequality constraints on the trajectory $h(\boldsymbol{\xi}(t)) = 0,\ g(\boldsymbol{\xi}(t)) \leq 0$, respectively. We define fixed start and goal configurations as equallity type of constraints, and joint limits as inequalities. We write them as inner product with kernel functions in the RKHS:\\ 
 \meq{}{
 h(\cdot)^\top\bs{y}&\leftarrow\langle \boldsymbol{\xi},K(t_o,\cdot)\bs{y}\rangle_{\mathcal{H}} - \boldsymbol{q_o}^\top\bs{y}=0,\ \boldsymbol{q_o}\in\mathcal{C},\ \text{for}\ t_o=\{0,1\} \\
 g(\cdot)^\top\bs{y}&\leftarrow\langle \boldsymbol{\xi},K(t_p,\cdot)\bs{y}\rangle_{\mathcal{H}}-\boldsymbol{q_p}^\top\bs{y}\leq0,\ \boldsymbol{q_p}\in\mathcal{C},\ \text{for}\ t_p=[0,1] 
 }
 for any $\bs{y}\in \mathcal{C}$, writting each configuration as the 
the respective Lagrange multipliers, $\bs\gamma^o,\ \bs\mu^p \in \mathbb{R}^{D}$ to the objective function \eref{eq:mini}, associated with each constraint $o$, $p$, yields:
 \meq{}{\label{eq:constsol}
 \boldsymbol{\xi}^{n+1}(\cdot)=&\ \text{arg}\min\limits_{\boldsymbol{\xi}}\ \ \langle \boldsymbol{\xi}-\boldsymbol{\xi}^n, \nabla \mathcal{U}[\boldsymbol{\xi}^n] \rangle_{\mathcal{H}} + \frac{\lambda}{2}\|\boldsymbol{\xi}-\boldsymbol{\xi}^n\|^2_{\mathcal{H}}  + {\bs{\gamma^o}}^\top h[\bs\xi] + {\bs{\mu^p}}^\top g[\bs\xi]
 }

 Solving the KKT system for the stationary point of \eref{eq:constsol} for ($\boldsymbol{\xi}, \gamma^o,\mu^p$), with $\mu^p\geq 0$, we obtain the constrained solution \eref{eq:constsolution}. \\Let $ \textrm{d}c_j \equiv \bs{J}^\top(t_j,u_j)\nabla c\left(x(\boldsymbol{\xi}^n(t_j),u_j)\right)$.The full update rule becomes:
 \meq{}{\label{eq:constsolution}
 \boldsymbol{\xi}^{*}(\cdot) = &\left(1-\frac{\beta}{\lambda}\right)\boldsymbol{\xi}^n(\cdot) -\frac{1}{\lambda}\left( \sum_{t_j\in \boldsymbol{\mathcal{T}}}  K(t_j,\cdot)\textrm{d}c_j + K(t_o,\cdot)\bs{\gamma^o} + K(t_p,\cdot)\bs{\mu^p}\right)
 }
 This constrained optimization solution, ends up augmenting the finite support set ($\mathcal{T}$) with points that are in constraint violation, weighting the kernel functions by the respective Lagrange multipliers. Each of the multipliers can be interpreted as a quantification of how much the points $t_o$ or $t_p$ violate the respective constraints.

\end{document}